\documentclass[letterpaper, 10 pt, journal]{IEEEtran} 
\IEEEoverridecommandlockouts

\usepackage{cite}
\usepackage{amsmath,amssymb,amsfonts}
\usepackage{graphicx}
\usepackage{textcomp}
\usepackage{gensymb}
\usepackage{comment}
\def\BibTeX{{\rm B\kern-.05em{\sc i\kern-.025em b}\kern-.08em
    T\kern-.1667em\lower.7ex\hbox{E}\kern-.125emX}}
    
\usepackage{algorithm}
\usepackage{algpseudocode}
\makeatletter
\def\BState{\State\hskip-\ALG@thistlm}
\makeatother
\usepackage[bottom]{footmisc}
\usepackage{xcolor}
\usepackage{cellspace} 
\usepackage{makecell} 
\setcellgapes{2pt}
\usepackage{booktabs}
\usepackage{hyperref}

\usepackage{caption}
\usepackage{subcaption}
\usepackage{tikz}

\begin{document}

\title{\LARGE{
Fast Functionally Redundant Inverse Kinematics for Robotic Toolpath Optimisation in Manufacturing Tasks
}}

\author{Andrew~Razjigaev, Hans~Lohr, Alejandro~Vargas-Uscategui, Peter~C.~King, and Tirthankar~Bandyopadhyay
	\thanks{This work was supported by CSIRO’s Early Research Career Postdoctoral and Engineering Fellowship. Financial support from CSIRO's Future Digital Manufacturing Fund is also gratefully acknowledged.}
	\thanks{A. Razjigaev, H. Lohr, A. Vargas-Uscategui, P. King and T. Bandyopadhyay are with the CSIRO, Australia. {\tt\small andrew.razjigaev@csiro.au}} 
}

\maketitle

\begin{abstract}
Industrial automation with six-axis robotic arms is critical for many manufacturing tasks, including welding and additive manufacturing applications; however, many of these operations are functionally redundant due to the symmetrical tool axis, which effectively makes the operation a five-axis task. Exploiting this redundancy is crucial for achieving the desired workspace and dexterity required for the feasibility and optimisation of toolpath planning. Inverse kinematics algorithms can solve this in a fast, reactive framework, but these techniques are underutilised over the more computationally expensive offline planning methods. We propose a novel algorithm to solve functionally redundant inverse kinematics for robotic manipulation utilising a task space decomposition approach, the damped least-squares method and Halley's method to achieve fast and robust solutions with reduced joint motion. We evaluate our methodology in the case of toolpath optimisation in a cold spray coating application on a non-planar surface. The functionally redundant inverse kinematics algorithm can quickly solve motion plans that minimise joint motion, expanding the feasible operating space of the complex toolpath. We validate our approach on an industrial ABB manipulator and cold-spray gun executing the computed toolpath. 




\end{abstract}

\begin{IEEEkeywords}
Robot Kinematics, Redundant Manipulators, Functional Redundancy, Toolpath Optimisation, Additive Manufacturing
\end{IEEEkeywords}

\begin{figure}[htbp]
\centering
     \begin{subfigure}[b]{0.4\textwidth}
         \centering
       \includegraphics[width=\textwidth]{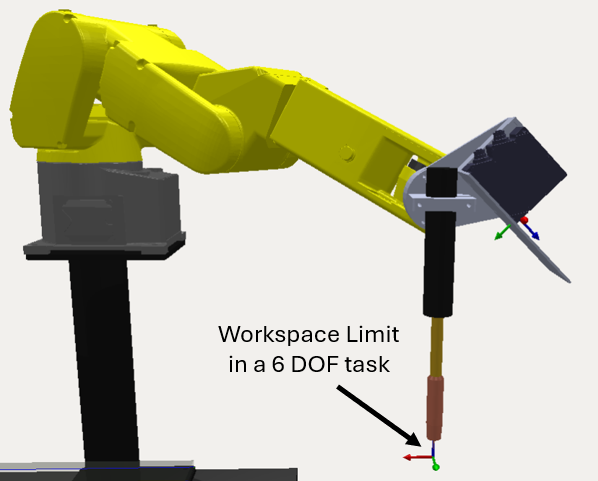}
         \caption{Workspace limit in a 6 DOF task}
         \label{fig:ik}
     \end{subfigure}
     \hfill
     \begin{subfigure}[b]{0.4\textwidth}
         \centering
         \includegraphics[width=\textwidth]{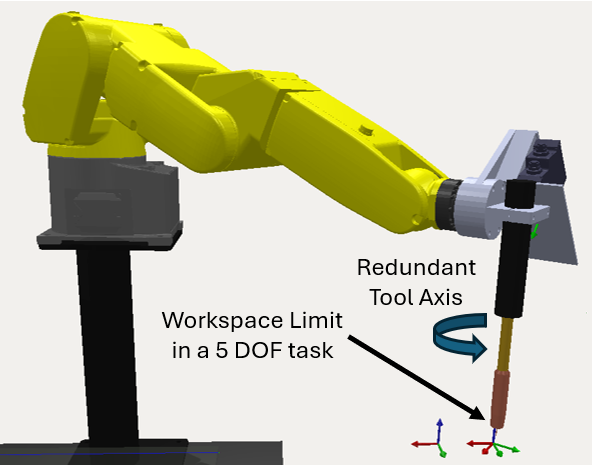}
         \caption{Workspace limit in a 5 DOF task}
         \label{fig:frik}
     \end{subfigure}

        \caption{Comparison of the maximum reach of a welding tool moving along a line in the x-axis on a six-axis robot when constrained with a 6 DOF task versus a 5 DOF task}
        \label{fig:FRFig}
\end{figure}

\section{Introduction}
Industrial automation solutions in manufacturing rely on six-axis robotic manipulators to accurately position tools and sensors to perform various tasks. The applications of six-axis manipulators include machining, painting, spraying, welding, grinding, laser-cutting, and additive manufacturing (AM), where the robot uses its 6 degrees of freedom (DOF) to move tools attached to its end-effector along a predefined path that is generally tangent to the surface of a workpiece. 

Traditional industrial robot motion planning requires practitioners to define toolpaths in a 3D Cartesian operation space of 6 DOF; however, the practical operation of these tools does not require constraints in all DOF. The most notable case is the symmetrical tool axis, where rotation about that axis is irrelevant to the task. For instance, the cutting profile of a tool in machining is symmetrical, as is the shape of the deposition profile of AM processes such as welding in wire-arc AM (WAAM) or in coating and spraying processes such as cold spray AM. 

Consequently, toolpaths for these applications only need constraints in 5 DOF, introducing kinematic redundancy where the six-axis manipulator has a single DOF greater than the number of variables that are necessary to describe the given task. The literature on robot kinematics \cite{Rigelsford2000ModellingAC} describes this type of redundancy as functional redundancy $r < m$, as the task has $r = 5$ components in the operation space of $m = 6$, despite that the robot is fully actuated in joint space, i.e. $n = m = 6$. In terms of motion planning with functionally redundant manipulators, solving the path in the operation space given the reduced constraints in the task is a type of task-constrained planning often described as a semi-constrained or an under-defined Cartesian path planning problem \cite{edwards2015descartes}. 

Motion planning with kinematic redundancy can provide greater dexterity and versatility to a manipulator in a number of different situations \cite{Rigelsford2000ModellingAC}. We illustrate these benefits in Fig. \ref{fig:FRFig} where a six-axis Fanuc LRmate200ic robot, is tasked with moving a weld torch along a line on the x-axis in our robot modelling and toolpath generation software called Continuous3D\textsuperscript{TM} \cite{Continuous3D, Lohr2023Continuous3DAE}. Continuous3D\textsuperscript{TM} is software made by CSIRO that has been designed for robotic additive manufacturing and repair techniques. For example, Lomo et al. recently used Continuous3D\textsuperscript{TM} to 3D print a titanium-based engine bracket by cold spray additive manufacturing \cite{LOMO202312, LOMO2023103891}. In coupling toolpath generation with robot motion planing, Continuous3D\textsuperscript{TM} aims to provide a more optimum solution for the execution of additive toolpaths robotically. Defining the task in 6 DOF as in \ref{fig:ik} imposes an unnecessary constraint on the manipulator that reduces its reachable workspace. In contrast, defining the task in 5 DOF as in \ref{fig:frik} allows the robot to use the redundant DOF to achieve a more reachable workspace with more intuitive movement. Exploiting functional redundancy can help optimise toolpath planning as it increases the set of feasible configurations, allowing motion planners to optimise trajectories for feasibility and performance by avoiding collisions, singularities, and minimising the motion, torque and energy involved in the task. However, despite these benefits, the literature seldom utilises inverse kinematics solvers for functionally redundant robots, especially in the domain of additive manufacturing where smooth continuous end effector motions have to be strictly adhered to.

\subsection{Related Work}
The literature on functionally redundant robot motion either resolves the redundancy in the path planning step or in the inverse kinematics step. The distinction is that the path planning approach solves the under-defined variables in the task such that $r$ becomes fully defined in $m$. On the other hand, the inverse kinematics approach employs techniques in the solver such that the effective operation space $m$ reduces to match the size of $r$. Both approaches have their benefits; the path planning approach produces a task that is executable by the traditional fully actuated inverse kinematics controller, but it is an offline method that requires pre-computation of the trajectory beforehand. While the inverse kinematics approach requires a controller customised to the task $r$, but it is an online method that can be executed in real-time, suitable for reactive motion control applications.

Amongst the path planning approaches in the literature, solving the under-defined variables in the task is a type of task-constrained motion planning and is solved with sampling and graph-based methods. For instance, the Descartes semi-constrained Cartesian path planning library by the ROS Industrial consortium would sample 6 DOF poses within the constraints of the functionally redundant task and build a collision-free graph in configuration space where the solver searches for the best path that minimises joint motion \cite{edwards2015descartes}. This approach also allows the task to have a tolerance of admissible values; for example, representing acceptable deviation of the tool approach angle as a cone of admissible orientations all of which can further increase the feasibility and smoothness of the robot motion. Sun et al. incorporated these techniques in the case of coordinated motion with external axes \cite{Sun2020semi}. Malhan et al. discuss the use of tolerance cones for orientation, spheres for position, as well as considering dynamically changing tool centre points (TCP), all within their path planner \cite{Malhan2023GenerationOC}. Sampling methods in functionally redundant motion have also been used to inform slicing algorithms in additive manufacturing to solve kinematically preferred rastering directions \cite{Bi2021GenerationOE}. However, an evaluation of graph-based techniques, particularly Descartes, has proven to be computationally expensive at large toolpaths in terms of both computation time and memory requirements \cite{Maeyer2017descartesEval}. Weingartshofer et al. proposed an optimisation framework for solving semi-constrained paths, however, they are also computationally expensive \cite{WEINGARTSHOFER2023opt}. Path planning solvers for finding motion for functionally redundant manipulators are highly flexible and practical, but can suffer due to computationally complexity in larger toolpaths. 

As opposed to planning, functionally redundant motion can be accounted for in the inverse kinematics solver, particularly using redundancy resolution techniques. A classic approach to solve functionally redundant inverse kinematics is to add virtual joints to the end of the kinematic model, where the under-defined variables in the task become the values of the virtual joint positions \cite{baron2000jointlimits}. This approach converts the functionally redundant problem into an intrinsically redundant one, increasing the size of $n$ and shifting the difference between $r$ and $m$ towards a difference between $m$ and $n$. In this case, task priority redundancy resolution techniques are employed \cite{Nakamura1987taskprio}, where the inverse kinematics solver can fulfil the primary task of reaching the target while utilising the Null Space for secondary objectives, which is the set of redundant self-motion configurations that do not change the TCP location \cite{Rigelsford2000ModellingAC}. In the context of functional redundancy, the Null Space also includes the set of configurations that fulfil all $r$ components in the task. Baron et al. used this method to achieve joint limit avoidance as the secondary objective \cite{baron2000jointlimits}. However, adding virtual joints increases the robot DOF $n$, which can complicate the inverse kinematics process, increasing the size and ill-conditioning of the Jacobian matrix. To counteract this, research has focused on decomposition techniques that eliminate the components in $m$ and reduce them to $r$, resulting in a smaller Jacobian of size $r$ by $n$. Huo et al. proposed a twist decomposition approach (TWA) that can reduce the size of the Jacobian based on an orthogonal decomposition of the desired twist within the task priority inverse kinematics algorithm \cite{Huo2005KINEMATICIO}. Huo et al. applied this decomposition with redundancy resolution to solve functionally redundant motion in a welding task with secondary objectives that avoided joint limits and singularities, which improved its overall motion along a toolpath \cite{Huo2008TheJA, Huo2011TheSO}. However, Zlajpah et al. proved that the formulation of TWA was mathematically inconsistent, proposing the consistent task-space decomposition (TSD) method \cite{lajpah2021ATS}, based on their formulation of a kinematic velocity controller for a changing task space \cite{lajpah2017changing, lajpah2019UnifiedVG}. Zlajpah et al. demonstrated that functionally redundant inverse kinematics can be solved efficiently in a real-time controller; however, these fundamental techniques are still niche and underutilised in the literature.  

\subsection{Contribution}
Exploiting redundancy is a key technique in optimising toolpaths for kinematic performance and feasibility, especially in functionally redundant tasks that are under-constrained in operation space. However, the literature predominantly resolves this with offline planning methods, seldom utilising the fast online control methods. Online control methods can potentially produce effective toolpaths with minimum computation, which can be essential for real-time control applications and in solving more computationally complex planning problems. One example is multi-levelled optimisation of slicing parameters and workpiece position with semi-constrained paths for kinematic performance in AM. 

In this paper, we propose a novel algorithm to solve functionally redundant inverse kinematics that is fast and simple to compute, robust to singularities and minimises the joint motion in fulfilling the task. Our algorithm achieves these qualities by incorporating task space decomposition, damped least squares (Levenberg-Marquardt optimisation) and Halley's method. We demonstrate the applicability of our algorithm in optimising toolpaths to reduce robot motion and increase the feasible operating space, where a robot is tasked with performing a complex non-planar toolpath in cold spray AM. 

\section{Problem Formulation}
In our manufacturing application, a toolpath is generated using slicing algorithms in our software, Continuous3D\textsuperscript{TM} \cite{Continuous3D, Lohr2023Continuous3DAE}. This toolpath is the desired trajectory of the TCP throughout the duration of the task. We can represent this trajectory as a discrete array of $N$ target poses relative to the robot's origin.

\begin{equation}
    \mathbf{T}_d (k) =  
\begin{bmatrix}
\mathbf{R}_d(k) & \mathbf{t}_d(k) \\
\mathbf{0}^\top & 1
\end{bmatrix}, \quad k = 0, 1, \ldots, N
\end{equation}

Where $\mathbf{T}_d$ is a 3D homogeneous transform containing a rotation matrix $\mathbf{R}_d \in SO(3)$ and a translation vector $\mathbf{t}_d \in \mathbb{R}^3$ at any time step $k$. The rotation matrix encapsulates the normal vector of the task, where the z-axis of the TCP coordinate frame must align. Since the task is only concerned with this alignment, we have 5 DOF functional redundancy as the tool is unconstrained to rotate about this normal vector. 

Initially, this rotation is an arbitrary value, and the inverse kinematics solver must find a joint trajectory that fulfils the task, minimising its motion. For each sample $k$, the functionally redundant inverse kinematics solver must solve the joint position vector $\mathbf{q}$, which represents the configuration of $n$ DOF on the robot.

\begin{equation}
    \mathbf{q} = [q_1, q_2, \ldots, q_n]^T
\end{equation}

With the inverse kinematics solution of $\mathbf{q}$, we can compute the new TCP trajectory in our 6 DOF operating space using the forward kinematics, $f(\mathbf{q})$, to solve the end-effector pose $\mathbf{T}_e$.

\begin{equation}
    \mathbf{T}_e = f(\mathbf{q}) = \mathbf{T}_1(q_1) \mathbf{T}_2(q_2) \cdots \mathbf{T}_n(q_n)
\end{equation}

The following sections shall describe the method of solving the inverse kinematics for $\mathbf{q}$ as well as how to decompose the task as a 5 DOF problem.

\section{Damped Least Squares Inverse Kinematics}
The general approach to solving inverse kinematics in a fully actuated 3D operating space $r = m = 6$ is to use the resolved-motion rate-control scheme, utilising the instantaneous relationship between the end-effector and joint velocities.

\begin{equation}
  d\mathbf{x} = \mathbf{J} d\mathbf{q}
\end{equation}

Where end-effector 6 DOF  position and orientation velocities, $d\mathbf{x} \in \mathbb{R}^{m=6}$, is a $6\times1$ vector and the joint velocities, $d\mathbf{q} \in \mathbb{R}^n$, is a $n\times1$ vector. $\mathbf{J}$ is the manipulator Geometric Jacobian matrix of size $6 \times n$ that contains the partial derivatives relating $d\mathbf{q}$ and $d\mathbf{x}$. The inverse kinematics is solved iteratively with the Newton-Raphson method designed to reduce the error $\mathbf{e}$ to zero, which is a $6 \times 1$ vector, a spatial twist between the end-effector pose $\mathbf{T}_e$ and the desired target pose $\mathbf{T}_d$ \cite{quik2022}.

\begin{equation}
    \mathbf{e} = \log(\mathbf{T}_e \mathbf{T}^{-1}_d)
\end{equation}

To converge the error, $d\mathbf{x}$ is set in the direction of $\mathbf{e}$; however, at large magnitudes, the iterative steps become inaccurate, so the magnitude is usually saturated by a value $e_{max}$.

\begin{equation}
    d\mathbf{x} = 
\begin{cases}
\mathbf{e}, & \text{if } \|\mathbf{e}\| \leq e_{max} \\
e_{max} \cdot \frac{\mathbf{e}}{\|\mathbf{e}\|}, & \text{if } \|\mathbf{e}\| > e_{max}
\end{cases}
\end{equation}

We can then use $d\mathbf{x}$ to solve the joint update $d\mathbf{q}$ by computing the pseudo-inverse of the Jacobian matrix $n \times 6$.

\begin{equation}\label{IK}
  d\mathbf{q} = \mathbf{J}^{\dagger} d\mathbf{x}
\end{equation}

The Moore-Penrose expression of the pseudo-inverse, $\mathbf{J}^{\dagger} =  \mathbf{J}^T (\mathbf{J} \mathbf{J}^T)^{-1}$ provides a minimum-norm solution minimising the residual error $\|d\mathbf{x} - \mathbf{J} d\mathbf{q}\|^2$. However, when the pseudoinverse is evaluated near, but not exactly at a singularity, it can result in a poorly behaved solution, giving large commanded joint angles and destabilising the iterative algorithm \cite{Deo1995OverviewOD, quik2022}. Conversely, the damped least-squares approach (Levenburg-Marquardt optimisation) can resolve this issue as it minimises both the residual error and the joint update $\|d\mathbf{x} - \mathbf{J} d\mathbf{q}\|^2 + \lambda\|d\mathbf{q}\|^2$ where $\lambda$ is a damping factor used to regularise convergence. This gives us a singularity robust pseudo-inverse \cite{Deo1995OverviewOD}.

\begin{equation}
    \mathbf{J}^{\dagger} = 
    \mathbf{J}^T (\mathbf{J} \mathbf{J}^T + \lambda^2 \mathbf{I})^{-1}
\end{equation}

Overall, for each $i^{th}$ iteration of the solver, the update is computed in Eq. \ref{DampedLeastSquares} with the damped least squares pseudo-inverse. 

\begin{equation}\label{DampedLeastSquares}
d\mathbf{q} = \mathbf{J}^T(\mathbf{JJ}^T + \lambda^2 \mathbf{I})^{-1} d\mathbf{x}
\end{equation}

The joint positions $\mathbf{q}$ iteratively converge to the inverse kinematics solution until the error $\mathbf{e}$ converges towards zero.

\begin{equation}
    \mathbf{q}_{i+1} = \mathbf{q}_i + d\mathbf{q}
\end{equation}

The following section describes how damped least squares changes in formulation with the task-space decomposition to achieve functional redundancy.

\section{Task Space Decomposition Approach}

In functional redundancy, we must decompose and reduce both $\mathbf{J}$ and $d\mathbf{x}$ into the new task of $r$ components. This changes the formulation of Eq. \ref{IK} with the modified Jacobian $\mathbf{\hat{J}}$, a $r \times n$ matrix and $d\mathbf{\hat{x}}$, a $r \times 1$ vector.

\begin{equation}\label{FunIK}
d\mathbf{q} = \mathbf{\hat{J}}^{\dagger} d\mathbf{\hat{x}}
\end{equation}

To compute $\mathbf{\hat{J}}$ and $d\mathbf{\hat{x}}$, task-space decomposition requires them to be transformed with respect to the desired target frame $\mathbf{T}_d$ \cite{lajpah2017changing, lajpah2021ATS}. Given that the velocities in $\mathbf{J}$ and $d\mathbf{x}$ are with respect to the origin, the twist transformation, $\mathbf{\tilde{R}}_t$, is a $6 \times 6$ matrix that transforms them to be with respect to the desired target frame.

\begin{equation}\label{rotationtransform}
\mathbf{\tilde{R}}_t =     
\begin{bmatrix}
        \mathbf{R}_d & \mathbf{0}_{3\times3}\\
        \mathbf{0}_{3\times3} & \mathbf{R}_d
\end{bmatrix}   
\end{equation}

Where $\mathbf{R}_d$ is the rotation matrix of the desired target frame relative to the robot's origin. With the quantities expressed relative to the target frame, a task projection matrix $\mathbf{T}_t$ performs the decomposition and reduction to tasks $r$ to the Jacobian and twist.

\begin{equation}\label{FunJ}
\mathbf{\hat{J}} = \mathbf{T}_t \mathbf{\tilde{R}}_t \mathbf{J}
\end{equation}

\begin{equation}\label{Fundx}
d\mathbf{\hat{x}} = \mathbf{T}_t \mathbf{\tilde{R}}_t d\mathbf{x}
\end{equation}

The task projector matrix, $\mathbf{T}_t$ is a mask computed as a sub-matrix (of size $r \times 6$) of the identity $\mathbf{I}_{6\times6}$ depending on components selected in $r$. For the case of achieving a 5 DOF task, $\mathbf{T}_t = \mathbf{T}_{r=5}$ eliminates the rotation about the z-axis, making $\mathbf{\hat{J}}$ a $5 \times n$ and $d\mathbf{\hat{x}}$ a $5 \times 1$ vector.

\begin{equation}\label{FunTt5}
\mathbf{T}_{r=5} =
\begin{bmatrix}
        \mathbf{I}_{5\times5} & \mathbf{0}_{5\times1}
\end{bmatrix} 
\end{equation}

In the case of achieving a 3 DOF task where the robot is only concerned of achieving the position, $\mathbf{T}_r = \mathbf{T}_{r=3}$ eliminates all orientation, making $\mathbf{\hat{J}}$ a $3 \times n$ and $d\mathbf{\hat{x}}$ a $3 \times 1$.

\begin{equation}\label{FunTt3}
\mathbf{T}_{r=3} =
\begin{bmatrix}
        \mathbf{I}_{3\times3} & \mathbf{0}_{3\times3}
\end{bmatrix} 
\end{equation}

Consequently, the variables in Eq. \ref{FunJ} and \ref{Fundx} can be substituted into Eq. \ref{DampedLeastSquares} to compute the inverse kinematics solution, which minimises $\|d\mathbf{\hat{x}} - \mathbf{\hat{J}} d\mathbf{q}\|^2 + \lambda\|d\mathbf{q}\|^2$.

\begin{equation}\label{DampedLeastSquaresFun}
d\mathbf{q} = \mathbf{\hat{J}}^T(\mathbf{\hat{J}}\mathbf{\hat{J}}^T + \lambda^2 \mathbf{I})^{-1} d\mathbf{\hat{x}}
\end{equation}

The following section describes how to achieve faster convergence of this algorithm with Halley's method.

\section{Fast Convergence with Halley's Method}

Lloyd et al. proposed using Halley's method, a third-order algorithm that uses both the first- and second-order derivatives to iteratively solve inverse kinematics much faster than the classic Newton-Raphson approach \cite{quik2022}. This requires the computation of the kinematic Hessian $\mathbf{H} \in \mathbb{R}^{6 \times n \times n}$, a rank-3 tensor that is the derivative of the Jacobian $\mathbf{J}$.

\begin{equation}
    \mathbf{H} = \frac{\partial}{\partial \mathbf{q}} \mathbf{J}
\end{equation}

Lloyd et al. presented that $\mathbf{H}$ can be efficiently computed based on the variables in the Jacobian $\mathbf{J}$. The algorithm can also be damped, requiring two steps to combine the first- and second-order derivatives. The first step is the damped Newton-Raphson step producing the joint update $\partial \mathbf{q}_{dnr}$.

\begin{equation}\label{DampedLeastSquaresHalleys}
\partial \mathbf{q}_{dnr} = \mathbf{J}^T(\mathbf{JJ}^T + \lambda^2 \mathbf{I})^{-1} d\mathbf{x}
\end{equation}

The second step produces the damped Halley's step to compute the final joint update $d\mathbf{q}$ based on the augmented matrix $\mathbf{A}$, with the multiplication $[\mathbf{H} \partial\mathbf{q}_{dnr}] \in \mathbb{R}^{6 \times n}$.

\begin{equation}\label{DampedLeastSquaresHalleys2}
d\mathbf{q} = \mathbf{A}^T(\mathbf{AA}^T + \lambda^2 \mathbf{I})^{-1} d\mathbf{x}
\end{equation}

\begin{equation}\label{augmented}
    \mathbf{A} = \mathbf{J} + \frac{1}{2} \mathbf{H} \partial\mathbf{q}_{dnr}
\end{equation}

This method minimises $\|d\mathbf{x} - \mathbf{J} d\mathbf{q} - \frac{1}{2} \mathbf{H} d\mathbf{q} d\mathbf{q}\|^2 + \lambda\|d\mathbf{q}\|^2$, simultaneously solving the quadratic system while minimising the joint update $d\mathbf{q}$.

Applying the task-space decomposition with Halley's method requires transforming the accelerations in the kinematic Hessian with respect to the target frame, as well as decomposing it into $r$ tasks with the task projection matrix $\mathbf{T}_t$.

\begin{equation}\label{FunH}
\mathbf{\hat{H}} = \mathbf{T}_t \mathbf{\tilde{R}}_t \mathbf{H}
\end{equation}

Where $\mathbf{\hat{H}} \in \mathbb{R}^{r \times n \times n}$ is the modified Hessian for the task of $r$ components. Substituting Eq. \ref{FunJ} and \ref{FunH} into \ref{augmented} produces the modified augmented matrix $\mathbf{\hat{A}}$.

\begin{equation}\label{FunA1}
\mathbf{\hat{A}} = 
\mathbf{T}_t \mathbf{\tilde{R}}_t \mathbf{J} + \frac{1}{2} \mathbf{T}_t \mathbf{\tilde{R}}_t 
\mathbf{H} 
\partial\mathbf{q}
\end{equation}

\begin{equation}\label{FunA}
\mathbf{\hat{A}} = 
\mathbf{T}_t \mathbf{\tilde{R}}_t (\mathbf{J} + \frac{1}{2} 
\mathbf{H} 
\delta\mathbf{q})
= \mathbf{T}_t \mathbf{\tilde{R}}_t \mathbf{A}
\end{equation}

Consequently, Eq. \ref{FunA} shows that the modified augmented matrix can be computed the same way that the other modified terms are, as in Eq. \ref{FunJ}, and \ref{Fundx}. The resulting damped Newton-Raphson step with functional redundancy is then computed based on Eq. \ref{DampedLeastSquaresFun}.

\begin{equation}\label{FunIKhalley1}
\partial \mathbf{q}_{dnr} = \mathbf{\hat{J}}^T(\mathbf{\hat{J}\hat{J}}^T + \lambda^2 \mathbf{I})^{-1} d\mathbf{\hat{x}}
\end{equation}

The second damped Halley's step with functional redundancy is computed based on Eq. \ref{DampedLeastSquaresHalleys2} using the modified augmented matrix $\mathbf{\hat{A}}$ from Eq. \ref{FunA}.

\begin{equation}\label{FunIKhalley2}
d\mathbf{q} = \mathbf{\hat{A}}^T(\mathbf{\hat{A}\hat{A}}^T + \lambda^2 \mathbf{I})^{-1} d\mathbf{\hat{x}}
\end{equation}

The resulting joint update $d\mathbf{q}$ solves the functionally redundant problem quickly with reduced motion minimising the quantity $\|d\mathbf{\hat{x}} - \mathbf{\hat{J}} d\mathbf{q} - \frac{1}{2} \mathbf{\hat{H}} d\mathbf{q} d\mathbf{q}\|^2 + \lambda\|d\mathbf{q}\|^2$, simultaneously solving the quadratic system while minimising the joint update. Pseudo-code of the solver is presented in algorithm \ref{FRIKpseudo}.

\begin{algorithm}[H]
\caption{Functionally Redundant Inverse Kinematics (FRIK) algorithm}\label{FRIKpseudo}
\begin{algorithmic}[1]
\Require Desired end-effector pose $\mathbf{T}_d$, initial joint configuration $\mathbf{q}_0$, convergence tolerance $\epsilon$
\Ensure Joint configuration $\mathbf{q}$ such that the $r$-DOF task-space projection of $f(\mathbf{q})$ matches that of $\mathbf{T}_d$
\State $\mathbf{q} \gets \mathbf{q}_0$
\Repeat
    \State Compute current end-effector pose: $\mathbf{T}_e \gets f(\mathbf{q})$
    \State Compute pose error: $d\mathbf{x} \gets \log(\mathbf{T}_e \mathbf{T}_d^{-1})$
    \State Decompose into task space: $d\mathbf{\hat{x}} \gets \mathbf{T}_t \mathbf{\tilde{R}}_t \, d\mathbf{x}$
    \If{$\|d\mathbf{\hat{x}}\| < \epsilon$}
        \State \textbf{break}
    \EndIf
    \State Compute manipulator Jacobian: $\mathbf{J} \gets \dfrac{\partial f(\mathbf{q})}{\partial \mathbf{q}}$
    \State Project to task space: $\mathbf{\hat{J}} \gets \mathbf{T}_t \mathbf{\tilde{R}}_t \mathbf{J}$
    \State \textbf{(1) Damped Newton–Raphson step:}
        \[
        \partial \mathbf{q}_{dnr} \gets \mathbf{\hat{J}}^{T}
        (\mathbf{\hat{J}}\mathbf{\hat{J}}^{T} + \lambda^{2}\mathbf{I})^{-1} 
        d\mathbf{\hat{x}}
        \]
    \State Form augmented matrix: $\mathbf{A} \gets \mathbf{J} + \tfrac{1}{2}\mathbf{H}\,\partial\mathbf{q}_{dnr}$
    \State Project to task space: $\mathbf{\hat{A}} \gets \mathbf{T}_t \mathbf{\tilde{R}}_t \mathbf{A}$
    \State \textbf{(2) Damped Halley's step:}
        \[
        d\mathbf{q} \gets \mathbf{\hat{A}}^{T}
        (\mathbf{\hat{A}}\mathbf{\hat{A}}^{T} + \lambda^{2}\mathbf{I})^{-1} 
        d\mathbf{\hat{x}}
        \]
    \State Update joint configuration: $\mathbf{q} \gets \mathbf{q} + d\mathbf{q}$
\Until{maximum iterations reached or $\|d\mathbf{\hat{x}}\| < \epsilon$}
\State \Return $\mathbf{q}$
\end{algorithmic}
\end{algorithm}

\section{Experimental Validation: Toolpath Optimisation}
To test the proposed functionally redundant inverse kinematics (FRIK) solver in algorithm \ref{FRIKpseudo}, a simulation in Continuous3D\textsuperscript{TM} is set up to evaluate its effectiveness in improving the kinematic performance in executing a complex non-planar toolpath. As in our problem formulation, this toolpath initially has a manual ad hoc selection of the rotation about the symmetrical tool axis as needed in programming six-axis manipulators. We shall compare the robot motion plan in executing this toolpath with the motion plan generated by our FRIK algorithm, which will treat the symmetrical tool axis as a redundant DOF, i.e. a 5 DOF task, automatically selecting the appropriate rotation for that axis so that it minimises the robot motion.

\begin{figure}[htbp]
    \centering
    \includegraphics[width=0.9\linewidth]{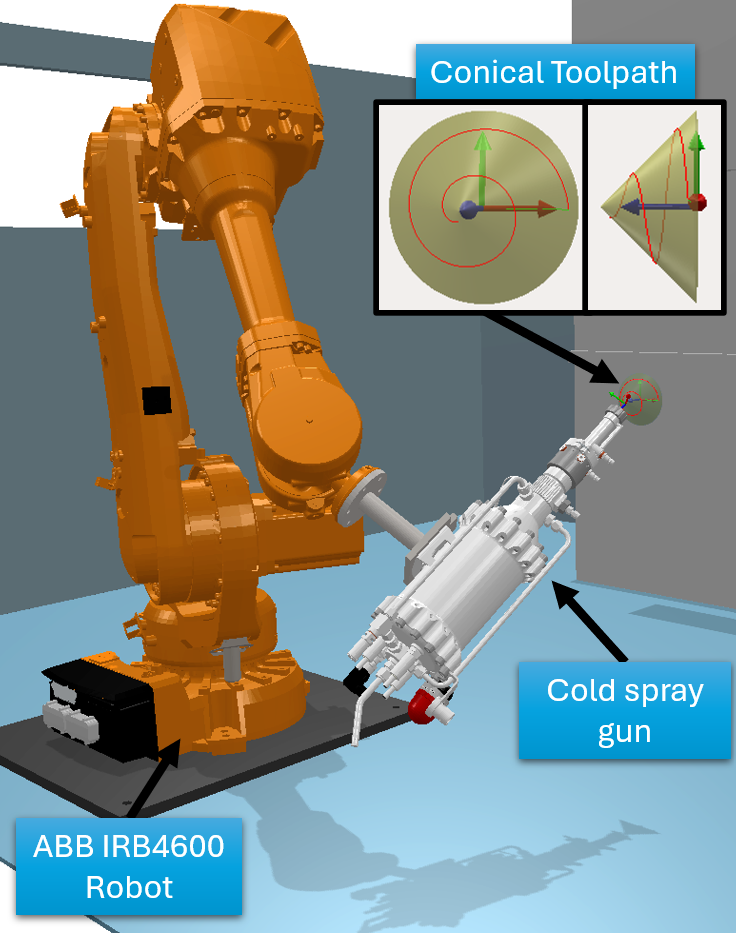}
    \caption{Experimental setup for simulation of the FRIK toolpath.}
    \label{fig:setup}
\end{figure}

The work-cell has an ABB IRB4600 robot holding an Impact Innovations GmbH 5/11 cold spray gun (Rattenkirchen, Germany) as its end-effector, as illustrated in Fig. \ref{fig:setup}. The Denavit-Hartenberg parameters of the robot are presented in Table \ref{table:DH}, which were automatically computed in Continuous3D\textsuperscript{TM} \cite{Lohr2023Continuous3DAE}. The robot assumes the starting configuration of $\mathbf{q}_0$.

\begin{equation}
    \mathbf{q}_0 = [-112\degree, -7\degree, 57\degree, -80\degree, -34\degree, 9\degree]^T
\end{equation}

\begin{table}[htbp]
\caption{Denavit-Hartenberg parameters}
\centering
{\makegapedcells
\begin{tabular}{c c c c c } 
\specialrule{.1em}{.05em}{.05em}
 DOF &  $a$ (mm) &$\alpha$ (radians) & $d$ (mm) & $\theta$ (radians) \\
 \hline
 1  &     175  &   $-\pi/2$     &  329.5    &       0\\
 2  &     900  &         0     &      0    & $-\pi/2$\\
 3  &  174.56  &   $-\pi/2$     &      0    &       0\\
 4  &       0  &   $-\pi/2$     &    960    & $\pi$\\
 5  &       0  &   $-\pi/2$     &      0    & $\pi$\\
 6  &       0  &         0     &       135 &       0\\
\specialrule{.1em}{.05em}{.05em}
\end{tabular}}
\label{table:DH}
\end{table}

The toolpath is a spiral coating trajectory on the surface of a cone (100mm in diameter and 50mm in height) generated by the non-planar slicing algorithm in Continuous3D\textsuperscript{TM} \cite{Lohr2023Continuous3DAE} as visualised in Fig. \ref{fig:setup}. The ad hoc selection of the rotation about the symmetrical tool axis is selected such that the x-axis stays aligned with the x-axis of the origin of the cone. Our FRIK solver, algorithm \ref{FRIKpseudo}, is developed in C++ with the Eigen linear algebra library.

In this experiment, we first find an appropriate placement of the workpiece (the cone) so that it is reachable in the robot's workspace and compare the feasible operating space between running the ad hoc and FRIK-designed toolpaths. Then, we assess the joint-space trajectory between the two motion plans. Lastly, we validate the trajectory by FRIK in the physical robot work-cell. 

\subsection{Workpiece Placement Analysis}
In the work cell, the placement of the workpiece is constrained by the surface of a wall at the base frame of the robot, specifically the -Y-Z plane, with the robot positioned as shown in Fig. \ref{fig:setup}. Dividing the plane into a grid of 100mm voxels allows us to assess whether the placement of the workpiece at each voxel results in a reachable toolpath. If so, we can determine the average manipulability in the motion plan and hence produce a map to find the feasible operating space. For this measurement, we take the joint-limit constrained Yoshikawa's manipulability index in Eq. \ref{jlmanipulability} based on \cite{Vahrenkamp2014RepresentingTR}.

\begin{equation}\label{jlmanipulability}
w_{\text{JL}}(\mathbf{q}) 
= \sqrt{\det\!\bigl( \mathbf{J} \mathbf{W} \mathbf{J}^T \bigr)}
\end{equation}

Where $\mathbf{W}$ is a diagonal weighting matrix of size $n \times n$ containing joint limit penalties $s_i(q_i)$ for each $i^{th}$ joint.

\begin{equation}
\mathbf{W}(\mathbf{q}) = \mathrm{diag}\!\left( s_1(q_1), \, s_2(q_2), \, \dots, \, s_n(q_n) \right)
\end{equation}

\begin{equation}
s_i(q_i) = \frac{(q_i^{\max} - q_i)(q_i - q_i^{\min})}{(q_i^{\max} - q_i^{\min})^2}
\end{equation}

\begin{figure}[t]
    \centering
    \includegraphics[width=\linewidth]{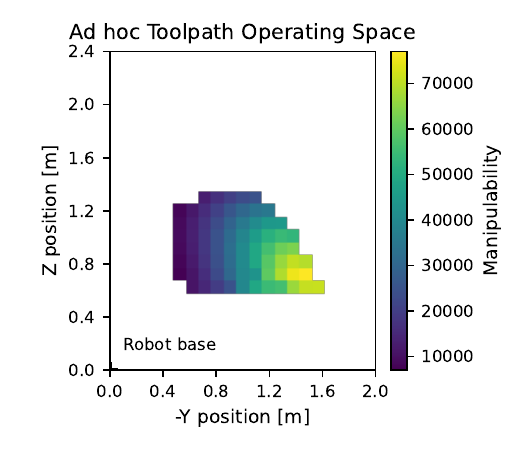}
    \caption{Feasible workpiece placement with ad hoc path}
    \label{fig:adhoc_workspace}
\end{figure}

\begin{figure}[t]
    \centering
    \includegraphics[width=\linewidth]{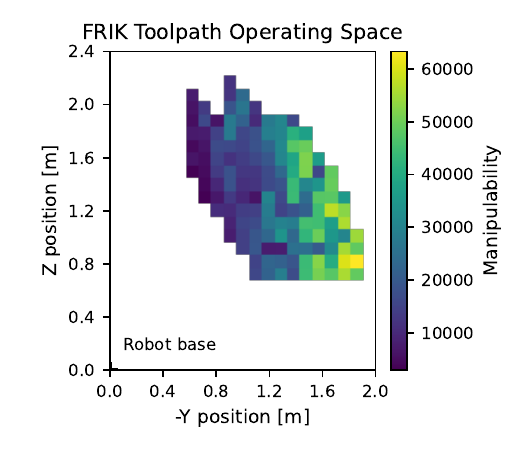}
    \caption{Feasible workpiece placement with FRIK path}
    \label{fig:frik_workspace}
\end{figure}

Figs. \ref{fig:adhoc_workspace} and \ref{fig:frik_workspace} illustrate the feasible operating space as heat maps between the toolpaths designed ad hoc and FRIK, respectively. The resulting feasible operating space for FRIK-designed toolpaths is visibly greater, emphasising how exploiting the redundant DOF helps the robot to achieve a more reachable workspace. Table \ref{table:workspace} compares the size and distribution of the ad hoc and FRIK toolpath operating spaces.

\begin{table}[t]
\caption{Workpiece Placement Analysis}
\centering
{\makegapedcells
\begin{tabular}{c c c c} 
\specialrule{.1em}{.05em}{.05em} 
 & Ad hoc & FRIK & \% Change\\
\hline
Reachable Voxels & 75 & 144 & 92\\
Maximum manipulability ($w_{\text{JL}}$) & 77,039.853 & 63,308.970 & -17.82\\
Mean manipulability ($\bar{w}_{\text{JL}}$) & 34,933.768 & 25,303.924 & -27.57\\
Std. dev. manipulability ($\sigma_{w_{\text{JL}}}$) & 20,160.141 & 15,443.855 & -23.39\\
\specialrule{.1em}{.05em}{.05em} 
\end{tabular}}
\label{table:workspace}
\end{table}

FRIK has a substantial 92\% increase in the number of reachable voxels compared to the ad hoc selection. FRIK also appears to spread the dexterity in the robot workspace; however, the peak of the distribution is lower than the ad hoc toolpath. The overall increase in workspace has not resulted in finding more dexterous positions, causing the mean manipulability and its standard deviation to decrease. This reduction in dexterity is a trade-off for increasing the reachable workspace. There are a few cases, near the base of the robot, where FRIK struggled to find a feasible motion, whereas the ad hoc selection was successful. This might reflect a limitation in FRIK, in that the inverse kinematics solver finds a locally optimal solution for instantaneous joint updates but may not achieve global optimality over long trajectories. Further work to improve our algorithm would involve integrating an initial search method to find the optimal $\mathbf{q}_0$ and solve the global trajectory with greater manipulability.

\subsection{Joint Trajectory Analysis}
\begin{figure}[t]
    \centering
    \includegraphics[width=\linewidth]{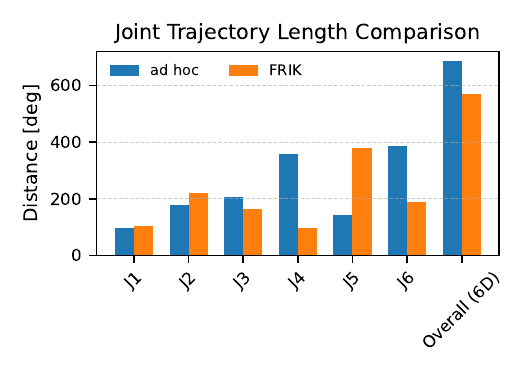}
    \caption{Joint Travel Comparison of FRIK and ad-hoc paths}
    \label{fig:jointtravel}
\end{figure}

\begin{figure}[htbp]
    \centering
    \includegraphics[width=0.9\linewidth]{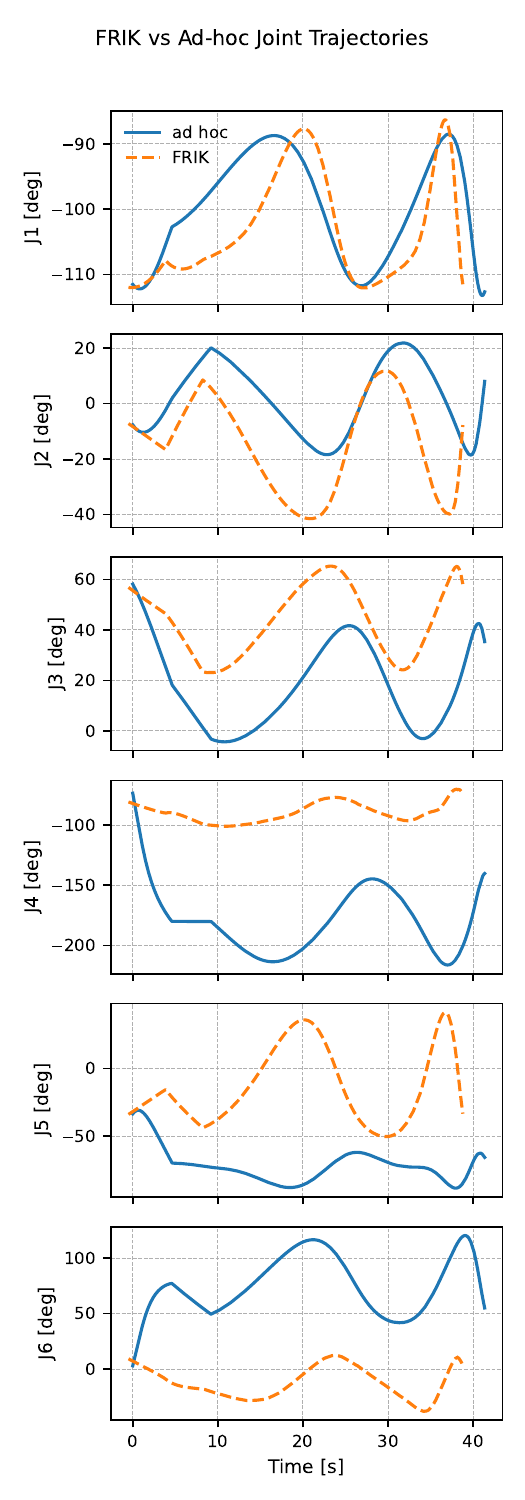}
    \caption{Joint Trajectory of the FRIK and ad-hoc paths}
    \label{fig:jointtraj}
\end{figure}

To assess the reduction in robot motion, we shall analyse the joint-space trajectory of $\mathbf{q}$ during the execution of the ad hoc and FRIK-designed toolpaths. The workpiece is kept at a constant position of $y=-1.1$m and $z=0.9$m. Figs. \ref{fig:jointtraj} illustrates the joint-space trajectory of all 6 DOF for both cases in degrees. Table \ref{table:joint} and Fig. \ref{fig:jointtravel} compare the overall distance traversed by each joint for both cases in degrees. As expected, FRIK has changed the distribution of motion along each joint, reducing joint numbers 3, 4 and 6 while concentrating more of the effort towards certain joint numbers 1, 2 and 5. This redistribution of joint motion resulted in decreasing the overall Euclidean distance traversed in joint space $\mathbf{q}$ by $16.66\%$. Therefore, FRIK changed the distribution of motion along each joint to achieve a more efficient trajectory in joint-space.

\begin{table}[hb]
\caption{Joint Travel Comparison of FRIK and ad-hoc paths}
\centering
{\makegapedcells
\begin{tabular}{c c c c} 
\specialrule{.1em}{.05em}{.05em} 
Joint    &         ad hoc [deg]   &     FRIK [deg]   &  \% Change\\
\hline
J1       &         95.711   &  102.347   &     6.93\\
J2       &        178.869   &  221.398   &    23.78\\
J3       &        206.085   &  165.112   &   -19.88\\
J4       &        356.713   &   97.687   &   -72.61\\
J5       &        141.283   &  379.122   &   168.34\\
J6       &        387.944   &  187.004   &   -51.80\\
\hline
Overall (6D)  &   685.549   &  571.313   &   -16.66\\
\specialrule{.1em}{.05em}{.05em} 
\end{tabular}}
\label{table:joint}
\end{table}

\subsection{Robot Validation}
\begin{figure}[t]
    \centering
    \includegraphics[width=1.5\linewidth, angle=90]{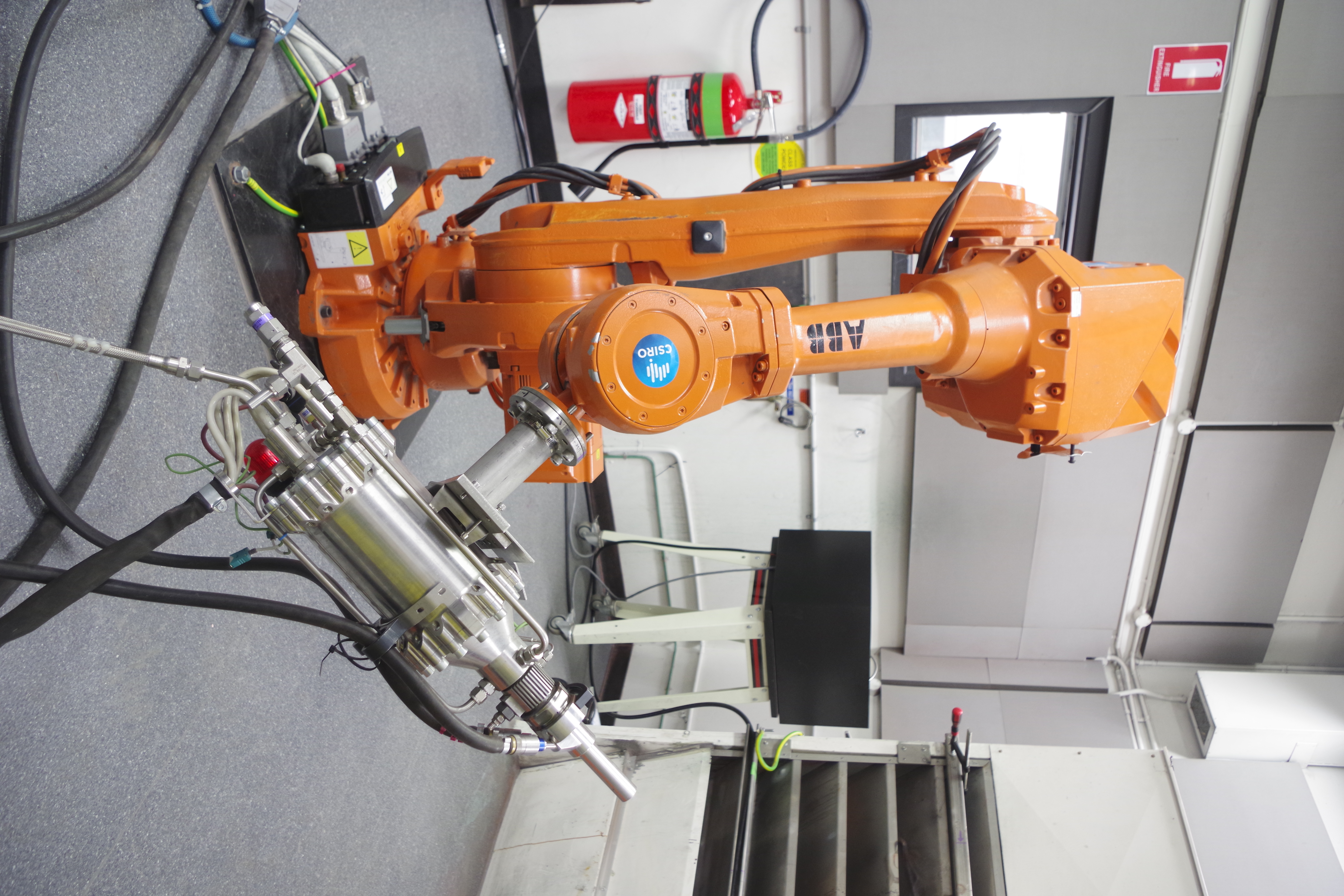}
    \caption{Experimental setup for robot validation of the FRIK toolpath.}
    \label{fig:lab}
\end{figure}

To validate the proposed method with real robotic hardware, the FRIK-designed conical toolpath is exported as a robot program, with Continuous3D\textsuperscript{TM}, that can be executed in ABB RobotStudio\textsuperscript{\textregistered} (Zurich, Switzerland) for our cold spray robotic work cell in CSIRO's Lab22 in Melbourne. Fig. \ref{fig:lab} shows the real setup of the robot in the lab. The FRIK algorithm was implemented in C++ and executed on a 13th Gen Intel i7 laptop with 32 GB RAM. The resulting processing time of the conical toolpath is presented in Table \ref{table:time}, each target pose was processed in approximately $200\mu s$ or $0.2ms$, suitable for real-time control. 



\begin{table}[t]
\caption{FRIK Computation Time in Conical Toolpath}
\centering
{\makegapedcells
\begin{tabular}{c c} 
\specialrule{.1em}{.05em}{.05em} 
 & Timed Results\\
\hline
Number of Linear Move Commands & 717\\
Simulation Steps ($N$) & 81,746\\
Average Processing time per step ($\mu$s) & 196.0\\
Total Processing Time (s) & 3.69\\
\specialrule{.1em}{.05em}{.05em} 
\end{tabular}}
\label{table:time}
\end{table}

\begin{figure*}[htbp]
    \centering
    \begin{subfigure}[b]{0.24\linewidth}
        \centering
        \includegraphics[width=\linewidth]{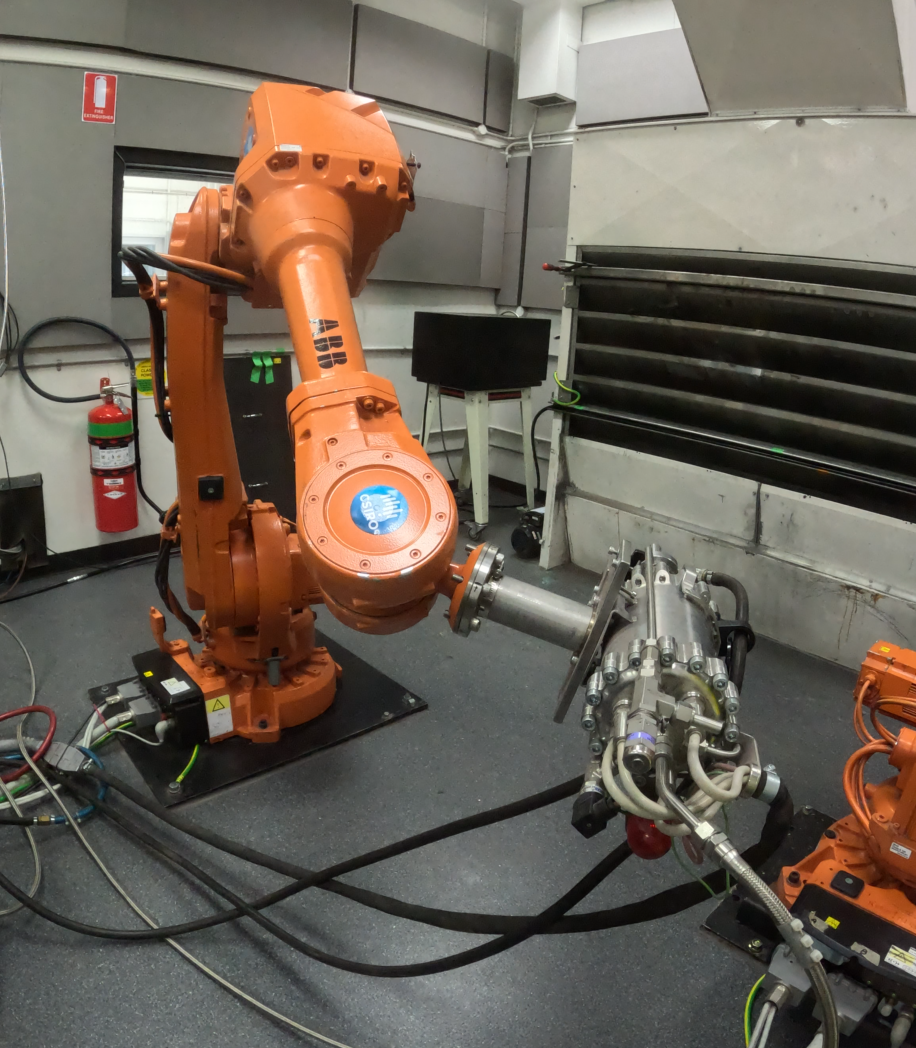}
        \caption{t = 10 s}
    \end{subfigure}
    \hfill
    \begin{subfigure}[b]{0.24\linewidth}
        \centering
        \includegraphics[width=\linewidth]{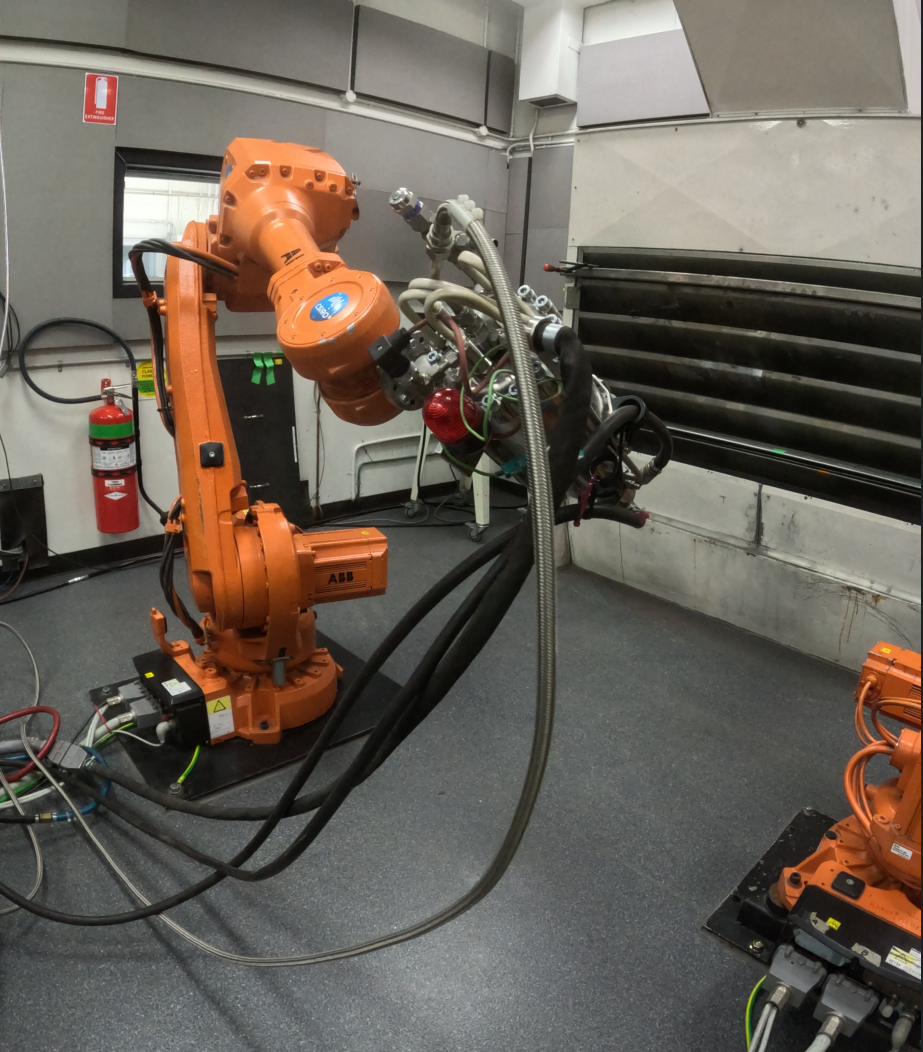}
        \caption{t = 16 s}
    \end{subfigure}
    \hfill
    \begin{subfigure}[b]{0.24\linewidth}
        \centering
        \includegraphics[width=\linewidth]{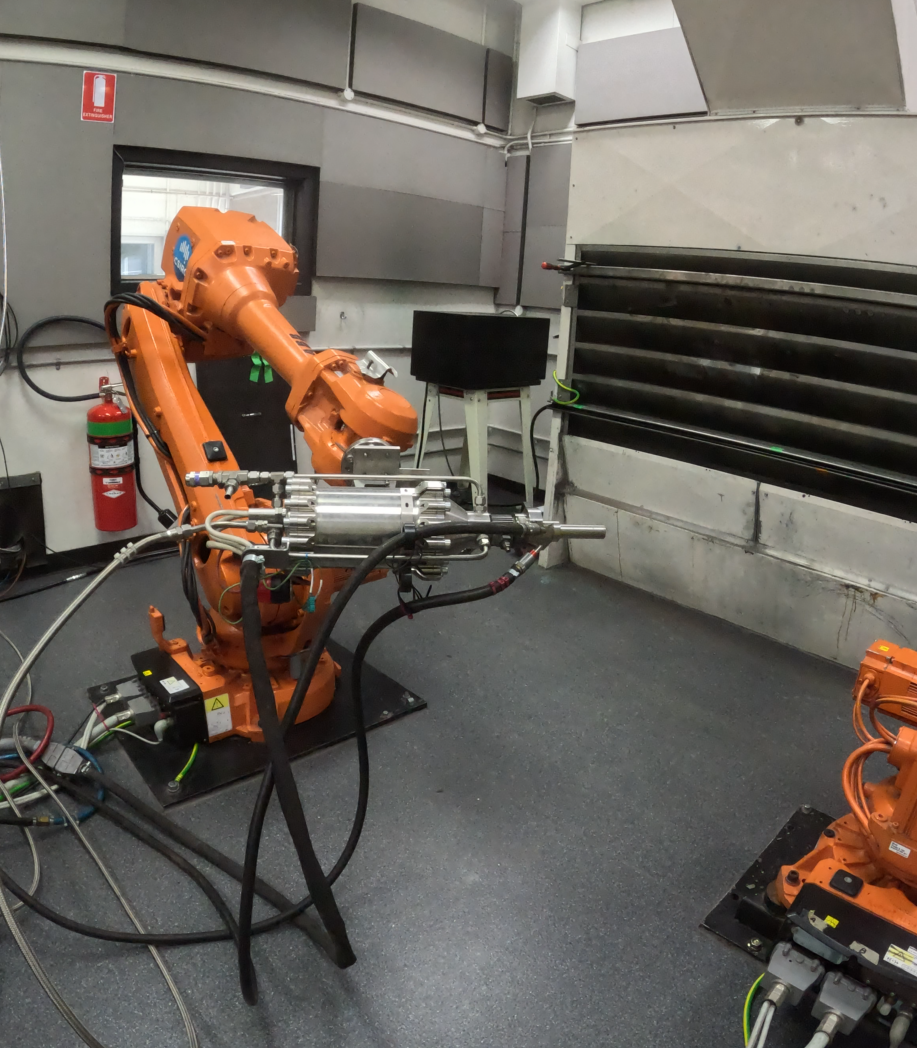}
        \caption{t = 23 s}
    \end{subfigure}
    \hfill
    \begin{subfigure}[b]{0.24\linewidth}
        \centering
        \includegraphics[width=\linewidth]{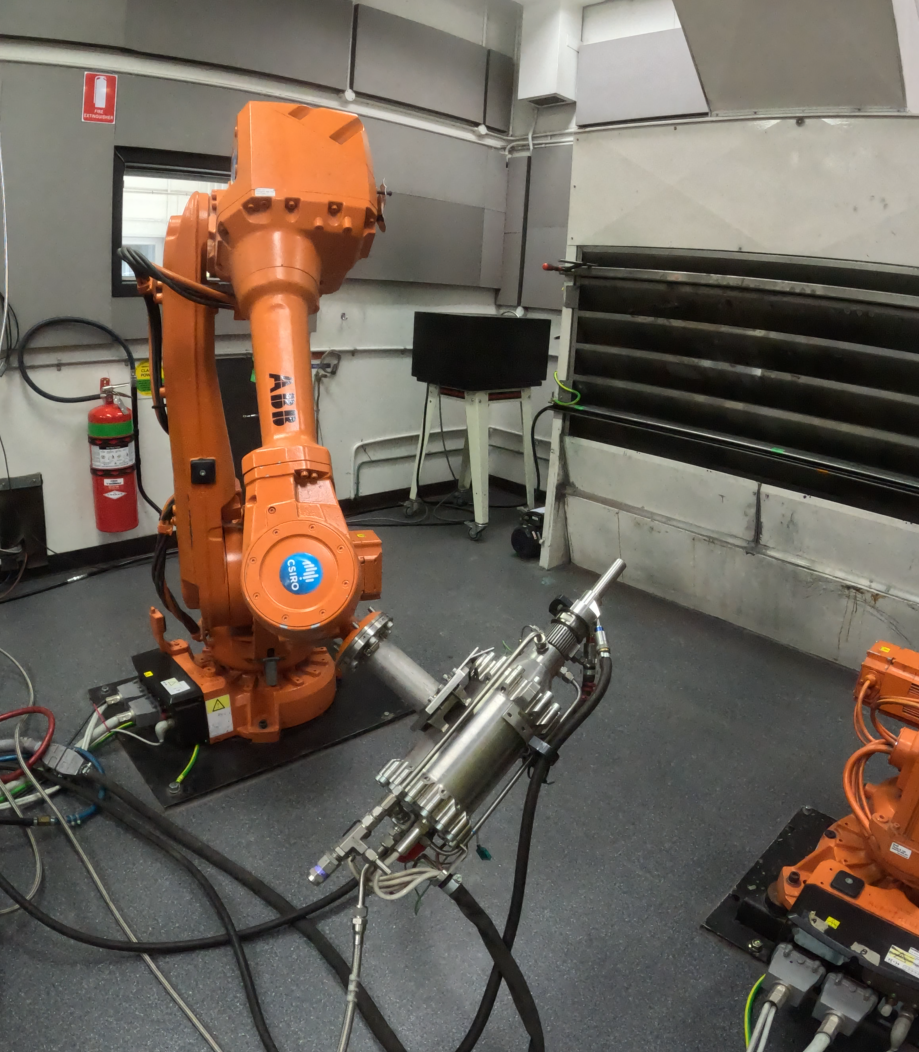}
        \caption{t = 29 s}
    \end{subfigure}
    
    \caption{Sequential frames extracted from robot validation experiment video of the FRIK planned conical toolpath.}
    \label{fig:video-sequence}
\end{figure*}

\begin{figure}[htbp]
    \centering
    \includegraphics[width=0.9\linewidth]{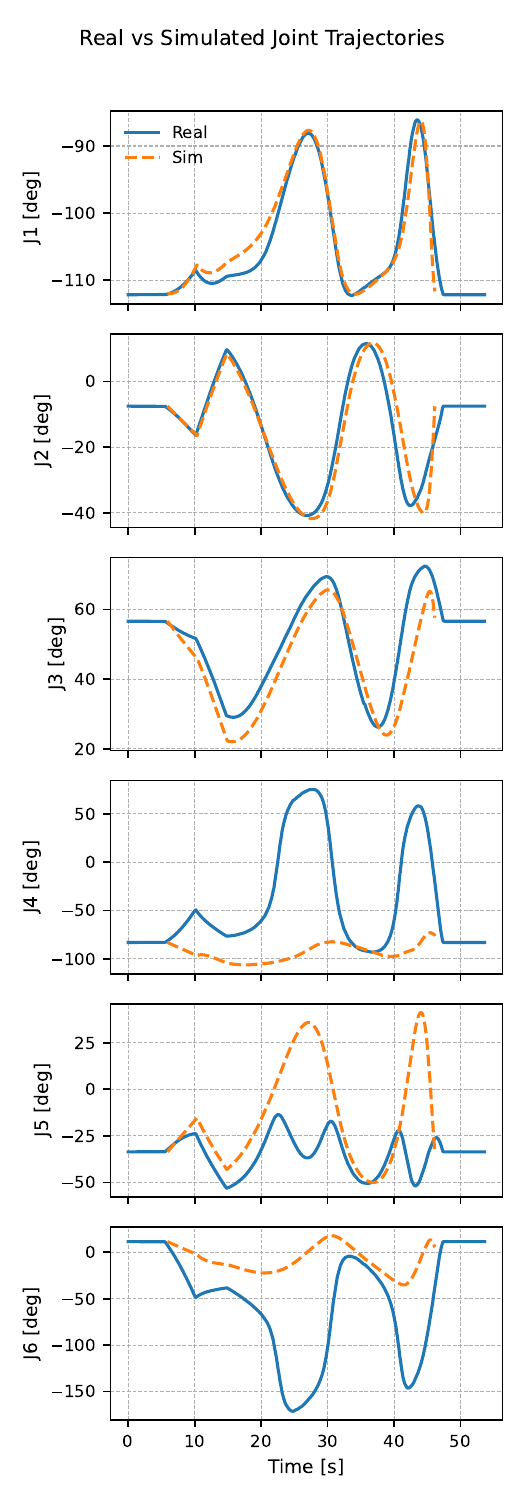}
    \caption{Joint Trajectory of the simulated and real robot}
    \label{fig:jointtrajreal}
\end{figure}

Fig. \ref{fig:video-sequence} shows frames extracted from the robot validation experiment video of the FRIK planned conical toolpath. Figs. \ref{fig:jointtrajreal} shows the data captured in RobotStudio\textsuperscript{\textregistered} tracking the joint trajectory during toolpath execution. The resulting trajectory demonstrates the behaviour as expected by the simulation; however, there are discrepancies in the joint trajectory, particularly with joints 4, 5 and 6. These discrepancies are due to the miscalibration of the location of the conical toolpath in the scene and the real lab setup. This highlights that kinematically-based toolpath processors, such as our FRIK algorithm, are susceptible to accuracy issues in setup in both simulated and real applications. Poor calibration can potentially compromise the benefits of using FRIK, which is a limitation of its usage. 

\section{Discussion}
The evaluated toolpath in this study was a non-planar spiral trajectory on the surface of a cone, which requires significant amounts of re-orientation of the tool, leaving only a small feasible operating space. The challenging toolpath makes it difficult for a practitioner to select the rotation about the redundant tool z-axis. Manual selection often results in an infeasible or inefficient motion plan. The resulting ad hoc selection results in a reduced operating space with difficulty in placing the workpiece, unintuitive robot movement, and placing significant emphasis on certain joints. Instead, allowing the redundant tool z-axis to be unconstrained, our FRIK algorithm was able to quickly and effectively exploit the redundant DOF, finding a reachable solution requiring minimum motion in terms of distance traversed in joint space. Exploiting the redundant symmetrical tool axis increases the reachable workspace by 92\% and reduces overall joint motion by 17\%. The results show that utilising FRIK in this toolpath helps to increase both the operating space of the toolpath and achieve more efficient robot movement. 

Although our FRIK algorithm has demonstrated effectiveness in finding feasible motion plans using a fast reactive inverse kinematics controller, a few limitations must be acknowledged. Firstly, FRIK with damped least squares is a local optimiser finding a solution that minimises the instantaneous joint update $d\mathbf{q}$ depending on the initial configuration $\mathbf{q}_0$. This greedy approach might not be globally optimal across large trajectories, which the more computationally intensive planning approaches would be able to solve as well as find the best $\mathbf{q}_0$. Also, while a damped FRIK can minimise the joint update and find a singularity robust solution, it does not necessarily use the redundancy for obstacle avoidance, joint-limit avoidance, torque optimisation or energy minimisation. These capabilities can be achieved in conjunction with the damped least squares as discussed in \cite{Deo1995OverviewOD}, but these secondary objectives are beyond the scope of this paper. When validating the robot motion plan in the lab, the miscalibration of the toolpath location resulted in affecting the real joint trajectory, highlighting the algorithm's sensitivity to the simulation-to-real gap. Lastly, the formulation of FRIK does not solve the more flexible, partially defined constraints described in semi-constrained path planning literature, such as the tolerance cone of admissible orientation. Future work may consider formulating FRIK to achieve these more flexible constraints and secondary objectives, as well as combinations with planning approaches to find globally optimal solutions.

\section{Conclusion}
In this paper, we develop a novel algorithm, FRIK, to solve functionally redundant inverse kinematics that is fast and simple to compute while minimising joint motion to fulfil the task. We used our algorithm to optimise the robot motion plan in executing a complex non-planar toolpath for a cold spray application where the z-axis of the tool is redundant and unconstrained. FRIK increased the feasible workspace by 92\% while reducing total joint-space traversal by 17\%, demonstrating its effectiveness in exploiting functional redundancy for complex toolpaths. In future work, we can consider improving our method by applying different secondary objectives and working with other types of constraints in semi-constrained paths. Further work would focus on addressing the sim-to-real gap by synchronising the scene in Continuous3D\textsuperscript{TM} with communication from the real robot. Combining the FRIK approach with sampling-based global planning can be an extension for finding globally optimal toolpaths.

\bibliographystyle{IEEEtran} 
\bibliography{references}


\end{document}